\pdfoutput=1

\documentclass[11pt]{article}

\usepackage{url}
\usepackage{graphicx}
\usepackage{wrapfig}
\usepackage{booktabs}
\usepackage{longtable}
\usepackage{tcolorbox}
\usepackage{stackengine}
\usepackage{multirow}
\usepackage{amsmath}
\usepackage{pifont}
\usepackage{floatflt}
\usepackage{algorithm}
\usepackage{algpseudocode}
\usepackage{booktabs,tabularx}

\usepackage{tcolorbox}
\usepackage[]{acl}

\usepackage{times}
\usepackage{latexsym}
\usepackage{amsmath}
\usepackage{graphicx}
\usepackage{booktabs}
\usepackage{bbm}
\usepackage{subcaption}
\usepackage{tabularx,ragged2e}
\usepackage{multicol,multirow}
\usepackage{enumitem}
\usepackage{soul}
\usepackage{tcolorbox}

\newcommand{\xmark}{\ding{55}}%
\newcommand{\cmark}{\ding{51}}%

\usepackage[T1]{fontenc}

\usepackage[utf8]{inputenc}

\usepackage{microtype}

\title{Power of Negation in Fostering LLM Hallucinations}
\title{Investigating and Addressing Hallucinations of LLMs in Tasks Involving Negation}

\author{Neeraj Varshney \hspace{14pt} Satyam Raj \hspace{14pt} Venkatesh Mishra \hspace{14pt}  Agneet Chatterjee \\ \textbf{Ritika Sarkar} \hspace{14pt} \textbf{Amir Saeidi} \hspace{14pt} \textbf{Chitta Baral}
  \\
  Arizona State University 
  }

\begin{document}
\maketitle

\begin{abstract}

Large Language Models (LLMs) have achieved remarkable performance across a wide variety of natural language tasks.
However, they have been shown to suffer from a critical limitation pertinent to `hallucination' in their output.
Recent research has focused on investigating and addressing this problem for a variety of tasks such as biography generation, question answering, abstractive summarization, and dialogue generation.
However, the crucial aspect pertaining to `negation' has remained considerably underexplored.
Negation is important because it adds depth and nuance to the understanding of language and is also crucial for logical reasoning and inference.
In this work, we address the above limitation and particularly focus on studying the impact of negation in LLM hallucinations.
Specifically, we study four tasks with negation: `false premise completion', `constrained fact generation', `multiple choice question answering', and `fact generation'.
We show that open-source state-of-the-art LLMs such as LLaMA-2-chat, Vicuna, and Orca-2 hallucinate considerably on all these tasks involving negation
which underlines a critical shortcoming of these models. 
Addressing this problem, we further study numerous strategies to mitigate these hallucinations and demonstrate their impact.

\end{abstract}

\section{Introduction}
\label{sec_introduction}

Despite the impressive performance achieved by recently developed Large Language Models \cite{touvron2023llama,NEURIPS2020_1457c0d6,chowdhery2022palm,rae2021scaling,smith2022using,mitra2023orca,vicuna2023}, their tendency to `hallucinate' in the output critically hampers their reliability and trustworthiness.
Hallucination in the LLM context corresponds to the generation of text that seems syntactically sound and correct but is factually incorrect or unfaithful to the source input \cite{Holtzman2020The,ji2023survey,maynez-etal-2020-faithfulness,zhang2023sirens}.

Prior work has studied hallucination of LLMs in various scenarios such as open-ended text generation \cite{manakul2023selfcheckgpt,varshney2023stitch}, question answering \cite{adlakha2023evaluating}, abstractive summarization \cite{chrysostomou2023lighter,aralikatte-etal-2021-focus,cao-etal-2022-hallucinated}, machine translation \cite{feng2020modeling}, and dialogue generation \cite{dziri-etal-2021-neural,sun2023contrastive}.
While the above studies are important, investigating the impact of `negation' in LLM hallucinations has remained underexplored.
Negation is important because it adds depth and nuance to the understanding of language.
It helps understand the opposite or absence of a statement, providing a more precise and nuanced interpretation and it is also crucial for logical reasoning and inference.
Furthermore, we humans arguably use affirmative expressions (without negation) more often than expressions with negation \cite{hossain-etal-2020-analysis,ettinger-2020-bert}; this implies that texts containing negation could be underrepresented in the training/tuning data of the models making it even more important to study.

\begin{figure*}[t!]
    \centering
    \includegraphics[width=14.4cm]{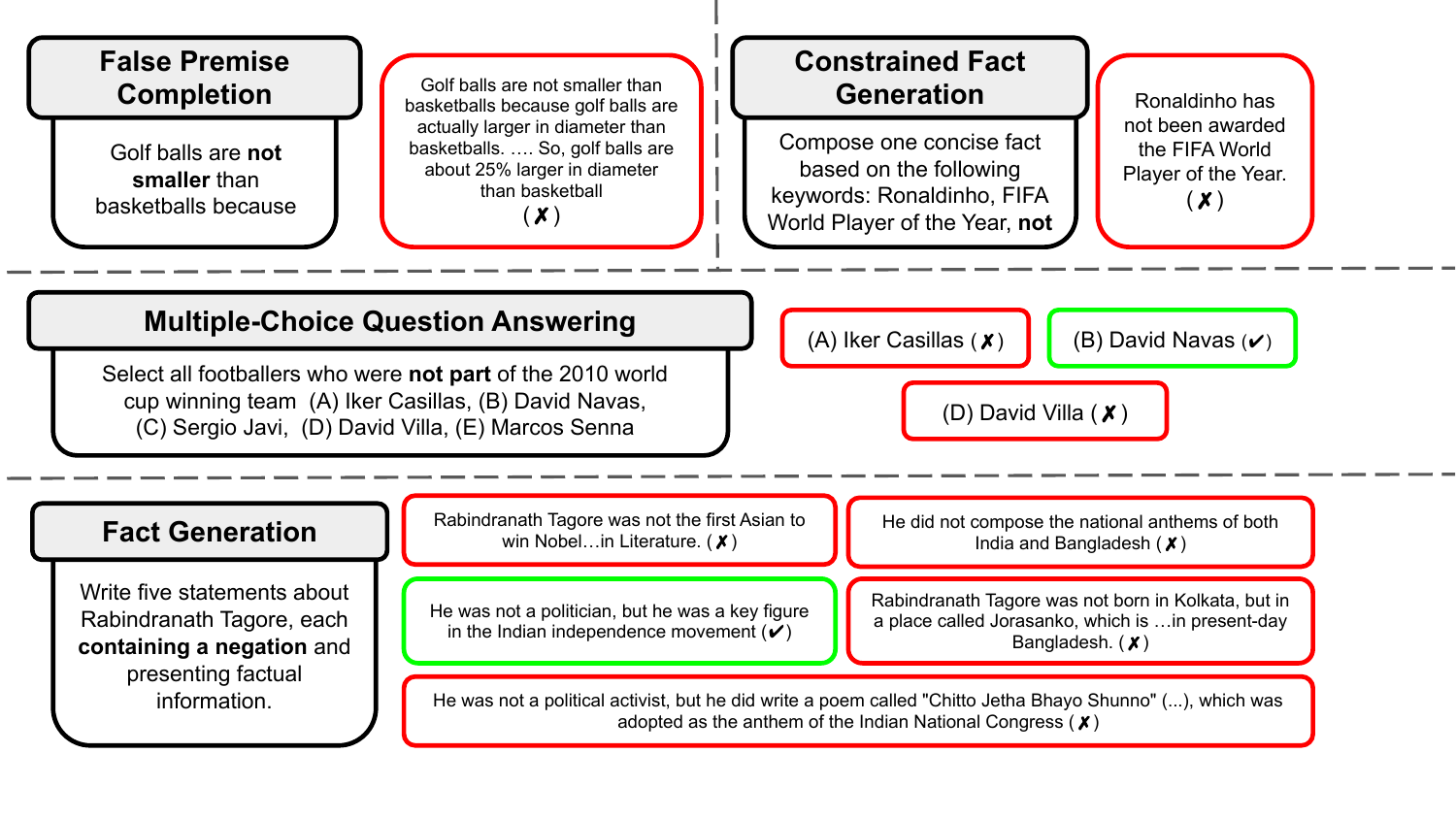}
    \caption{Illustration of the four tasks that deal with negation studied in this work. Responses enclosed in red boxes (marked with \xmark) are hallucinations while those in green boxes (marked with \cmark) are factually correct.}
    \label{fig:teaser}
\end{figure*}

With the aforementioned motivation, in this work, we focus on `negation' and study its impact on LLM hallucinations. 
Prior work on negation has primarily studied classification tasks such as natural language inference and masked word prediction \cite{hosseini-etal-2021-understanding,hossain-etal-2020-analysis,hossain-etal-2022-analysis,truong-etal-2023-language,kassner-schutze-2020-negated}. 
However, it is also important to study generative tasks with state-of-the-art LLMs.
To this end, we study negation in four tasks: 
(i) \textit{False Premise Completion} (FPC), 
(ii) \textit{Constrained Fact Generation} (CFG), 
(iii) \textit{Multiple-Choice Question Answering} (MCQA), and 
(iv) \textit{Fact Generation} (FG).
Figure \ref{fig:teaser} illustrates examples of all four tasks.
We provide a detailed description and the rationale behind studying these tasks in Section \ref{sec_evaluation_tasks}.

We comprehensively study the performance of various open-source state-of-the-art LLMs including LLaMA-2-chat \cite{touvron2023llama}, Vicuna-v1.5 \cite{vicuna2023}, and Orca-2 \cite{mitra2023orca}.
We show that these models hallucinate considerably on all the tasks.
On average, they hallucinate $63.77\%$, $72.33\%$, $36.6\%$, and $62.59\%$ on FPC, CFG, MCQA, and FG tasks respectively.
This underlines a critical limitation of these LLMs in effectively dealing with negation.

To address this hallucination problem, we further study various mitigation strategies such as providing a `cautionary instruction', demonstration via `in-context exemplars', `self-refinement' by leveraging the LLM's parametric knowledge, and `knowledge-augmented generation'.
Our study results in numerous important findings such as 
(a) providing a `cautionary instruction' along with `in-context exemplars' performs the best in mitigating the hallucinations though there remains a considerable room for improvement,
(b) providing contextual knowledge to the LLM when answering false premise prompts, coerces it to hallucinate even more instead of mitigation,
(c) `self-refinement' indeed mitigates the hallucinations to a certain extent; however, in some cases, it incorrectly transforms the output by introducing hallucinated information in the output.

Overall, our work underlines a critical shortcoming of existing LLMs and studies ways to mitigate it.
We will release our work to further facilitate future research in developing robust LLMs that can effectively deal with negation.

\section{Related Work}

Investigating the hallucination behavior of LLMs has attracted significant attention from the research community.
\citet{manakul2023selfcheckgpt, min-etal-2023-factscore, varshney2023stitch,dhuliawala2023chain} show that LLMs hallucinate when generating biography passages about various concepts.
\citet{jiang-etal-2023-active,kang2023ever} study multi-hop question answering using retrieval augmented generation. 
TruthfulQA \cite{lin-etal-2022-truthfulqa} focuses on evaluating the correctness of LLMs’ responses to questions.
There also exist discrimination based tasks such as HaluEval \cite{li2023helma} and FACTOR \cite{muhlgay2023generating} that focus on evaluating the ability to recognize hallucinations.
TruthfulQA \cite{lin-etal-2022-truthfulqa} also contains a discrimination format where it provides a multiple-choice alternative to test a model’s ability to identify truthful statements.
\citet{liu-etal-2022-token} focus on identifying conflicts in the context while \citet{lee2022factuality,muhlgay2023generating} directly prompt LLMs to complete text given a prefix.

We note that the above works investigating hallucinations lack comprehensively studying the crucial aspect of `negation'. 
In addition to the reasons mentioned in Section \ref{sec_introduction} for studying negation, 
we additionally note that negation also helps prevent misinterpretation of statements, i.e., without the ability to recognize negation, one might misunderstand the intended meaning of a sentence, leading to inaccurate responses.
In summary, negation is a fundamental aspect of linguistic expression and thus comprehensively studying it is important.

Prior studies on negation have primarily focused on classification tasks like natural language inference and masked word prediction.
\citet{hosseini-etal-2021-understanding} propose to fine-tune BERT \cite{devlin-etal-2019-bert} with an unlikelihood objective and evaluate on negated LAMA dataset \cite{kassner-schutze-2020-negated}.
\citet{hossain-etal-2020-analysis,hossain-blanco-2022-leveraging,truong-etal-2023-language} focus on natural language inference and cloze completion tasks with datasets like SNLI \cite{bowman-etal-2015-large}, Multi-NLI \cite{williams-etal-2018-broad}.
\citet{ye-etal-2023-assessing} have studied negation in logical reasoning context.
A more recent work \citet{jang2023can} study the performance of LLMs on transformed prompts of various datasets where the transformation is performed by replacing words like `correct' with `incorrect', `appropriate' with `inappropriate', and `natural' with `unnatural'.
This transformation results in prompts such as ``Complete the given sentence with the \textit{inappropriate} ending''.
Different from these studies, in our work, we focus on hallucinations of LLMs and conduct a systematic evaluation and analysis with four different generative tasks motivated from real-world settings.
Additionally, to address this hallucination problem, we also study various mitigation strategies.

\section{Evaluation Tasks}
\label{sec_evaluation_tasks}
In this section, we provide a detailed description
and the rationale behind studying all the tasks.

\subsection{False Premise Completion (FPC)}
This task consists of prompts that involve negation (not) and are based on false premises, i.e., incorrect presuppositions. 
We (the authors) first compile a list of fundamental facts from various domains such as Science, Geography, Sports, Animals, and Astronomy and then introduce a negation (not) while ensuring the grammatical correctness to create false premise prompts. 
Table \ref{tab:FPC_example_prompts} shows examples of this task and the distribution of prompts across the different domains.
For inference, we instruct the models to `complete the given prompt by providing factually correct information'.
Since the correct facts are negated, prompts in this task are factually incorrect; thus, a model needs to identify the false premise of the prompt and appropriately provide its response.

Consider a false premise prompt: ``Saturn is not the second largest planet in our solar system because'',
we show that models often falter on such false premise prompts and generate hallucinated responses such as ``\textit{ because it is actually the sixth largest planet in our solar system}''; however 
a robust model should respond to this false premise prompt with something like ``\textit{The statement in the prompt is incorrect because Saturn is indeed the second largest planet in our solar system, after Jupiter}''.
Note that we additionally study the performance on the corresponding correct premise prompts also as detailed in Section \ref{sec_results_FPC}.

\begin{table}[t]
    \small
    \centering
    \resizebox{\linewidth}{!}{
    \begin{tabular}{@{}p{0.2\linewidth}>{\RaggedRight}p{0.8\linewidth}@{}}
    \toprule
        \textbf{Domain} &
        \textbf{Prompts} 
        \\
    \toprule

        \multirow{4}{*}{\textbf{Science}} 
        & The speed of sound is \underline{not} affected by the medium through which it travels because\\
        \multirow{2}{*}{(39\%)} 
        & Heat energy does \underline{not} transfer from a warmer substance to a colder one because \\

        & Hydrogen does \underline{not} have atomic number of 1 because \\

        \midrule
        
        \multirow{3}{*}{\textbf{Astronomy}}  
        & Saturn is \underline{not} the second largest planet in our solar system because \\

        \multirow{1}{*}{(20\%)}  & Jupiter is \underline{not} bigger than Earth because \\
        
        \midrule
        
        \multirow{4}{*}{\textbf{Geography}}  
        & The Sahara Desert does \underline{not} have sand dunes because\\
        \multirow{2}{*}{(13\%)}  & The Arctic region does \underline{not} experience extreme cold temperatures because \\

        \midrule
        
        \multirow{2}{*}{\textbf{Animals} (8\%)}  
        & Chickens do \underline{not} lay eggs because \\

        & Tigers are \underline{not} carnivorous predators because \\

        \midrule
        
        \multirow{4}{*}{\textbf{Sports} (4\%)}  
        & India did \underline{not} win the 2011 world cup of cricket because \\
        & Golf balls are \underline{not} smaller than basketballs because \\

        \midrule
        
        \multirow{2}{*}{\textbf{Tech.} (3\%)}  
        & Floppy disks do \underline{not} have lower storage capacity than USB drives because \\

        \midrule

        \multirow{3}{*}{\textbf{Others} (9\%)}  
        & Inflation does \underline{not} decrease the purchasing power of money because \\

        & The square root of 64 is \underline{not} 8 because \\

    \bottomrule

    \end{tabular}
    }
    \caption{Examples of prompts for the FPC task.}
    \label{tab:FPC_example_prompts}
\end{table}

\paragraph{Rationale:}
We study this task because state-of-the-art models have been shown to perform well on a wide range of tasks that are based on correct presuppositions. 
However, users in real-world applications often tend to provide inputs that are based on false premises due to either the lack of relevant knowledge or to adversarially attack the system. 
Thus, the efficacy on this task is critical in preventing misinformation resulting from the hallucinated responses of the LLMs \cite{pan-etal-2023-risk}.
We attribute this kind of hallucination to the sycophantic behavior exhibited by LLMs \cite{sharma2023towards,ranaldi2023large}.

\begin{table}[t]
    \small
    \centering
    \resizebox{\linewidth}{!}{
    \begin{tabular}{@{}p{0.2\linewidth}>{\RaggedRight}p{0.8\linewidth}@{}}
    \toprule
        \textbf{Domain} &
        \textbf{Keywords} 
        \\
    \toprule

        \multirow{3}{*}{\textbf{Sports} (40\%)}        
        & Chris Froome, \underline{not}, Tour de France Winner \\

        & Sachin Tendulkar, \underline{not}, Cricket World Cup, 2011 \\

        & \underline{not}, Luka Modric, Ballon d'Or Winner \\

        \midrule
        
        \multirow{3}{*}{\textbf{Entertain}}  
        & Luke Combs, \underline{not}, Entertainer of the Year, CMA Awards \\

        \multirow{1}{*}{{(16\%)}}   & \underline{not}, Michael Jackson, Grammy Awards \\

        \midrule

        \multirow{1}{*}{\textbf{Award} (11\%)}  
         & \underline{not}, Ardem Patapoutian, Nobel Prize, 2021\\

        \midrule

        \multirow{2}{*}{\textbf{Politics} (13\%)}  
        & Barack Obama, US Presidential Election, \underline{not}, 2008 \\

        \midrule
        
        \multirow{2}{*}{\textbf{Others} (13\%)}  
        & The African Renaissance Monument, Senegal, tallest statue, \underline{not} \\

    \bottomrule

    \end{tabular}
    }
    \caption{Examples of keywords for the CFG task.}
    \label{tab:CFG_examples}
\end{table}

\subsection{Constrained Fact Generation (CFG)}
This task requires composing a fact based on the given keywords one of which is a negation (not).
Specifically, we use the following task instruction ``Compose one concise fact based on the following keywords''.
Note that despite the presence of `not' as a keyword, in all the instances of this task, there does indeed exist ways to compose factually correct responses from the provided keywords; however, a statement created by simply connecting `not' with the other keywords (in a syntactically sound manner) will result in a factually incorrect sentence.

Consider an example in which the keywords are ``The African Renaissance Monument, Senegal, tallest statue, not'',
simply creating a sentence by combining the keywords would result in ``The African Renaissance Monument statue in Senegal is not the tallest statue in Africa'' which is factually incorrect; 
however, a possible correct output is ``The African Renaissance Monument in Senegal, while being the tallest statue in Africa, is not the tallest statue in the world''.

Thus, it poses an important challenge for the models and requires true understanding of negation to compose a factually correct statement. 
Here, we focus on historical facts from the domains of Sports, Awards such as Nobel prizes, Politics, and Entertainment.
We particularly select these domains because information in these domains is unambiguously accurate and also easy to obtain and verify.
Table \ref{tab:CFG_examples} shows examples of this task.
Note that we also vary the position of `not' in the keyword list to avoid any bias in the models' outputs.

\begin{table}[t]
    \small
    \centering
    \resizebox{\linewidth}{!}{
    \begin{tabular}{@{}p{0.2\linewidth}>{\RaggedRight}p{0.8\linewidth}@{}}
    \toprule
        \textbf{Domain} &
        \textbf{Question} 
        \\
    \toprule

        \multirow{6}{*}{\textbf{Sports} (20\%)}  
        & Choose the countries that have \underline{not} hosted the Winter Olympics. 
        Options: Finland, Austria, China, South Korea, USA \\

        & Identify all the countries that have never played a FIFA World Cup Final.
        Options: Portugal, Belgium, USA,	Germany, Argentina \\

        \midrule
        
        \multirow{8}{*}{\textbf{Entertain}}  
        & Pick the musicians who have \underline{not} won a Grammy Award for Album of the Year. 
        Options: Babyface, John Mayer, Ed Sheeran,	Alanis Morissette, Taylor Swift \\

         \multirow{2}{*}{(12\%)}   & Identify the films that have \underline{not} won an Oscar for Best Film.
        Options: Anthony Adverse, The Irishman, Arrival	The Lord of the Rings: The Return of the King, All the King's Men. \\

        \midrule    
        
        \multirow{6}{*}{\textbf{Geo.} (27\%)}  
        & Identify all European cities that are \underline{not} capitals of their respective countries.
        Munich, Milan, Rome, Salzburg, Berlin \\

        & Identify all African countries from which the Nile does \underline{not} flow
        Options: Egypt, Burundi, Libya, Chad, Central African Republic \\
      
    \bottomrule

    \end{tabular}
    }
    \caption{Examples of questions for the MCQA task.}
    \label{tab:MCQA_examples}
\end{table}

\begin{table*}[]
    \centering
    \footnotesize
    \resizebox{\textwidth}{!}{
    \begin{tabular}{cccccc}
        \toprule
        \textbf{Sports} & \textbf{Politics} & \textbf{Music} & \textbf{Films and TV} & \textbf{Science} & \textbf{Literature} 
        \\
        \toprule
        Cristiano Ronaldo & Xi Jinping &  Michael Jackson &  Rihanna &  Albert Einstein & William Shakespeare \\
        Lionel Messi & Vladimir Putin &  The Beatles & Jackie Chan &  Marie Curie & Akira Toriyama \\
        Neymar Jr. & Donald Trump  &  Taylor Swift & Katy Perry &  Isaac Newton & Georges Simenon \\
        LeBron James & David Cameron &  Miley Cyrus & Deepika Padukone &  Galileo Galilei & Jin Yong \\ 
        Virat Kohli & Narendra Modi &  Justin Bieber & Jennifer Lopez &  Satyendra Nath Bose & J. K. Rowling \\
        \bottomrule
    \end{tabular}
    }
    \caption{Names of personalities from six distinct domains considered in the study for FG task.}
    \label{tab:internationl_personalities}
\end{table*}

\paragraph{Rationale:}
This task has numerous applications in information retrieval and search engines because generating facts based on keywords, even when negation is involved, enhances the effectiveness of search engines and is vital for users seeking precise, relevant, and accurate information in a vast sea of data.
This also has applications in automated content generation where users provide precise specifications to a generative system. 
It is also important to study this task for the prevention of misinformation from LLMs.

\subsection{Multiple-Choice QA (MCQA)}

In this task, a selection-based question involving negation is given along with multiple answer choices and the correct options that satisfy the question requirements need to be selected. 
Similar to the previous task, here, we focus on facts from the domains of Sports, Entertainment, Awards, etc. because these facts are unambiguously accurate and can be easily obtained and verified. 
Table \ref{tab:MCQA_examples} shows examples of this task.
Note that this is a multi-choice multi-correct QA task where multiple answer options can be correct.
In all the instances, we have a total of five answer options.

\paragraph{Rationale:}
This task is important in a variety of applications such as `medical diagnosis' where a system might encounter statements like ``the patient does not experience chest pain'' and it needs to rule out/select certain options by understanding the statement,
`legal document analysis' where the system can help quickly sift through clauses based on a given statement,
and
`customer service/sales chatbots' where sentences like ``I don't want red color t-shirts' are commonly encountered.

\subsection{Fact Generation (FG)}

This task requires generating statements about personalities, each containing a negation and presenting factual information.
To avoid any bias that may occur due to the lack of information, we include only widely known personalities.
Also, we select these personalities from diverse domains such as Sports, Politics, Music, Films \& TV, Science, and Literature.
Specifically, we select five personalities from each domain from the Forbes popular list as shown in Table \ref{tab:internationl_personalities}.

\paragraph{Rationale:}
This task is important in investigating misinformation which becomes very important when using LLMs to generate text about a person.
Moreover, in a general sense, while comparing different options in decision-making,  generating facts involving negation can help highlight the strengths and weaknesses of various options.

\section{Experiments and Results}

We experiment with various open-source state-of-the-art LLMs including LLaMA-2-chat \cite{touvron2023llama}, Vicuna \cite{vicuna2023}, and Orca-2 \cite{mitra2023orca}.
We experiment with the 13B parameter models and the evaluation set contains $300$ instances in FPC task ($150$ each of false premise and correct premise prompts as detailed in Section \ref{sec_results_FPC}), $100$ instances each in CFG and MCQA tasks, and $300$ instances ($5$ each of without negation and with negation for all the $30$ personalities) in FG task.
Note that all the tasks described in Section \ref{sec_evaluation_tasks} are focused on factuality; thus, we use the temperature of $0$ during inference.

\subsection{False Premise Completion}
\label{sec_results_FPC}

\paragraph{Performance Evaluation:} To quantify the performance, we measure the percentage of hallucinated responses.
Note that the performance evaluation on these tasks is challenging due to the existence of open-ended text responses to the inputs. 
Thus, we (the authors) perform human evaluations for this task which are more reliable. 
However, we also study LLM-based evaluations (using the Bard model) to bolster the human evaluations.

In human evaluations, we classify a response to one of the following three categories:
\textbf{Hallucinated (H)},
\textbf{Partially Hallucinated (PH)}, and
\textbf{Correct (C)}.
An `H' category response fails to identify the false premise of the prompt and provides hallucinated (incorrect) information.
A `PH' category response fails to identify the false premise of the prompt and completes it but provides correct information at some place in the response (by contradicting itself).
A `C' category response indeed detects the false premise of the prompt and provides the correct information by rectifying it.
Thus, `C' category response is the ideal response for this task over `PH' and `H'.
We note that this 3-level annotation is challenging to obtain from LLM-based evaluations; thus, we use only two levels (Correct and Incorrect) for LLM-based evaluations. 
We use the following instruction:
``Check if the given text is factually accurate based on the fact that `\{true fact\}'. Answer in `Correct' or `Incorrect'.
Text: \{model's output\}''.

\begin{table}[t]
    \resizebox{\linewidth}{!}{
    \begin{tabular}{p{1.5cm}p{1.7cm}p{2cm}p{1.5cm}}
    \toprule
        \textbf{Response} & \textbf{LLaMA-2}  & \textbf{Vicuna-v1.5} &  \textbf{Orca-2} \\
    \toprule
        
        \textbf{H} & 39.33\% & 58.67\% & 39.33\%\\
        \textbf{PH} & 14.00\% & 19.33\%  & 20.67\%\\
        \midrule
        \textbf{C ($\uparrow$)} & 46.67\% &  22.00\% & 40.00\% \\

    \bottomrule
    \end{tabular}
    }
    \caption{
    Performance of models on the FPC task.}
    \label{tab:FPC_model_comparison}
\end{table}

\begin{figure*}[t]
\centering
    \begin{subfigure}{.47\textwidth}
        \includegraphics[width=\linewidth]{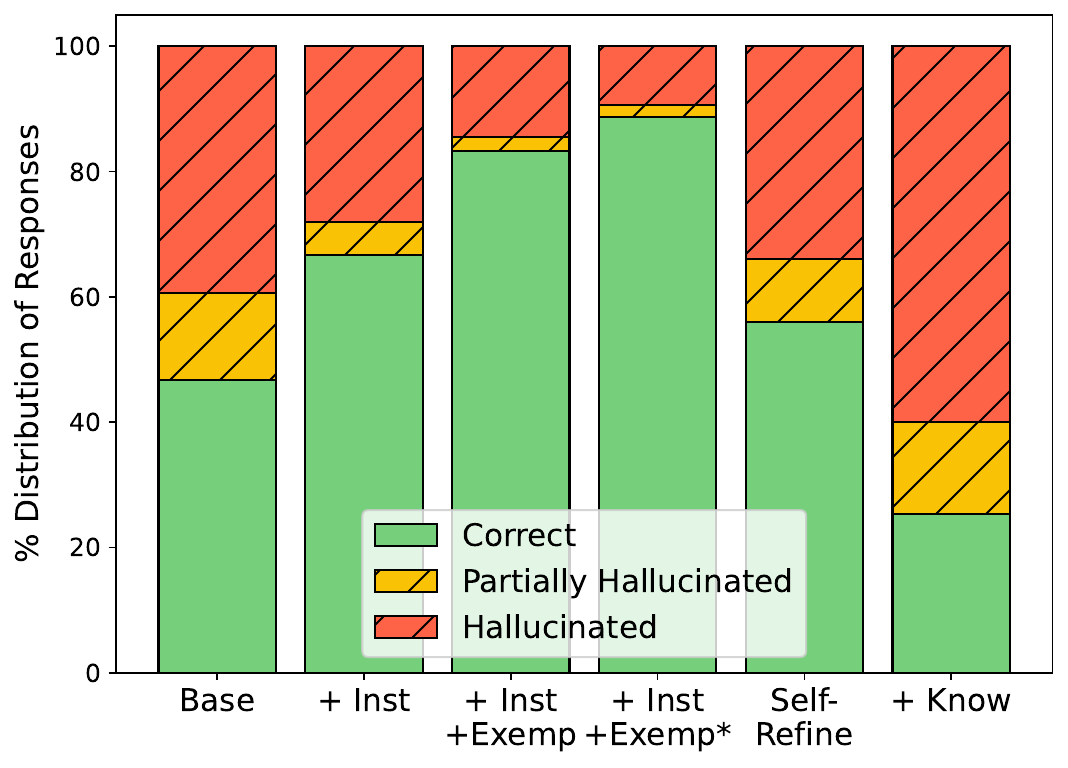}
    \caption{On False Premise Prompts}        
    \end{subfigure} 
    \begin{subfigure}{.47\textwidth}
        \includegraphics[width=\linewidth]{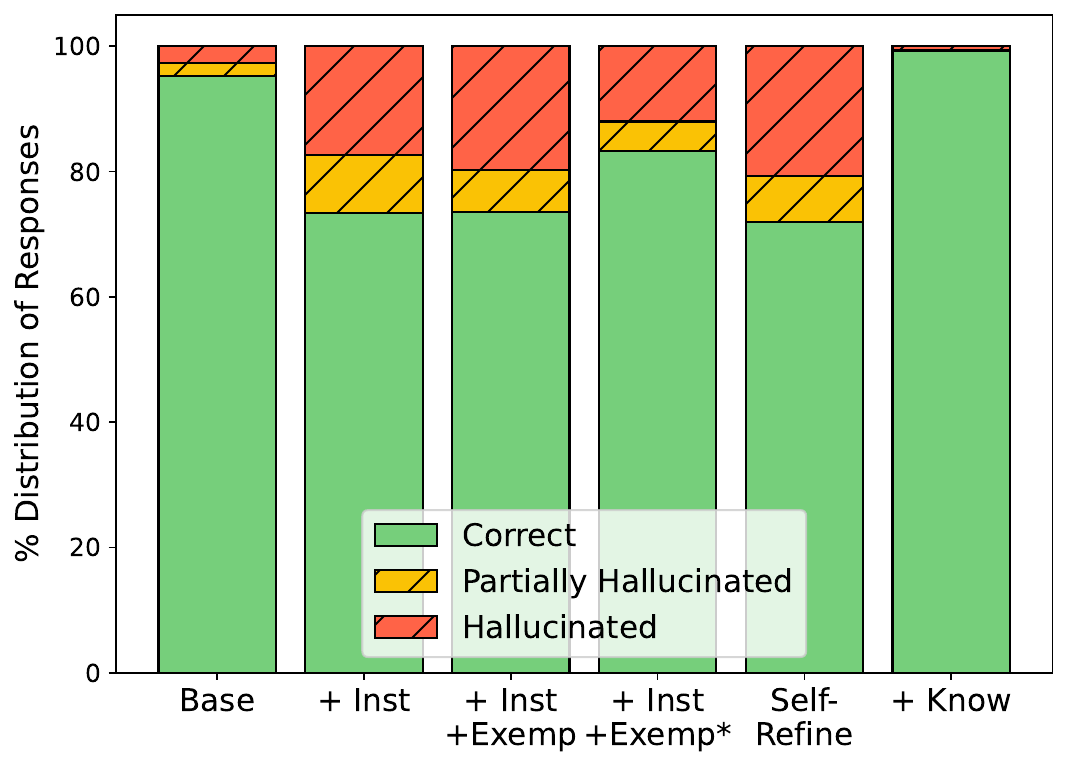}
    \caption{On Correct Premise Prompts}    
    \end{subfigure}
    
    \caption{Impact of various mitigation strategies with LLaMA-2 model on the Prompt Completion task. We show performance on both false premise prompts and correct premise prompts.}
    \label{fig:FPC_performance_both}    
\end{figure*}

\subsubsection{Performance of Models}
Table \ref{tab:FPC_model_comparison} shows the performance of various models on the FPC task.
Specifically, LLaMA-2, Vicuna-v1.5, and Orca-2 answer only $46.67\%$, $22\%$, and $40\%$ instances correctly with label `C'.
With BARD evaluations also, the correctness percentage is $44\%$.
It demonstrates that all the models hallucinate considerably on this task.
Table \ref{tab:FPC_example_responses} shows examples of responses of various models. 
We attribute this poor performance to the sycophantic behavior exhibited by the LLMs where they tend to generate responses that favor the user’s perspective present in the input rather than providing correct or truthful answers.
This result renders studying ways to mitigate these hallucinations important.

\subsubsection{Mitigation of Hallucinations}

To address the hallucination problem, we study a number of mitigation strategies on this task.
Note that the mitigation strategies can potentially deteriorate the performance on the `correct' premise prompts also. 
Thus, we also evaluate them on the \textbf{correct premise prompts}, i.e., the affirmative (non-negated) variants of the false premise prompts.
Table \ref{tab:FPC_correct_premise_prompts} shows examples of false premise prompts and corresponding correct premise prompts.

\paragraph{Cautionary Instruction (Inst):}
State-of-the-art LLMs have shown a remarkable ability to follow natural language instructions. 
Thus, a way to mitigate hallucination is by providing a cautionary instruction in the input.
Specifically, in addition to the task instruction (`Complete the given prompt by providing factually correct information'), we provide a cautionary instruction specifying `Note that the prompt can be misleading as well'.
We show that providing cautionary instruction indeed improves the performance to a certain extent.

\paragraph{Demonstrative Exemplars (Exemp):}
LLMs also possess an impressive ability to learn from demonstrations/exemplars provided in the input context.
To this end, we provide exemplars of input-output pairs of negation prompts in the context.
Note that we provide exemplars of negation prompts of both false premise and correct premise prompts so as to prevent biasing the models' outputs.
We experiment with three different combinations of exemplars (provided in the Appendix \ref{appendix_exemp}) and report the averaged results.
We note that there is no overlap between the evaluation instances and the demonstrative exemplars.

\paragraph{Self-Refinement (Self-Refine):}
In self-refinement strategy, we first obtain the model's output and then instruct it to `rewrite it by rectifying the factually incorrect information'.
This method attempts to leverage the parametric knowledge of the model in rectifying the potential mistakes in its output \cite{pan2023automatically}.

\paragraph{Knowledge Augmentation (Know):}
Here, we provide knowledge relevant to the prompt as additional contextual information to the LLM during generation. 
We use web search via Bing search API to obtain the relevant knowledge. 
Specifically, we use the input prompt as the query to retrieve the web search results. 
Appendix \ref{appendix_know} provides further details and examples of the retrieved knowledge.

\paragraph{Performance of Mitigation Strategies:}
Figure \ref{fig:FPC_performance_both}(a) shows the effectiveness of various mitigation strategies on the LLaMA-2 model's performance.
The bar corresponding to `Base' refers to the base setting without any mitigation strategy.
In `Inst' strategy, we add a cautionary instruction, and in `Inst + Exemp', we also add demonstrative exemplars.
`Inst + Exemp*' corresponds to the strategy where we provide exemplars of both negated and non-negated prompts (provided in Appendix \ref{appendix_exemp}).
The non-negated prompts exhibit just a slight impact on the false premise prompts; however, they play a crucial role on the correct premise prompts where we study the downside of these mitigation strategies (later in this Subsection).
We conduct additional analysis on a `self-checking' methodology in Appendix \ref{appendix_sec_fpc}.

It can be observed that all the strategies except `knowledge augmented generation' result in considerable improvements in reducing hallucinations. 
Table \ref{tab:FPC_flase_premise_mitigation_responses} shows examples of responses after application of various mitigation strategies on the false premise prompts.
We also analyzed the improvement of exemplars strategies and attribute their performance to the ability to counter the false premise prompt acquired from the in-context exemplars.
Also, we observe negligible deterioration (change from correct to incorrect) on the false premise prompts (except `Know' strategy) due to the mitigation strategies.

\paragraph{Knowledge coerces hallucination on false premise prompts:}
Knowledge considerably increases the hallucination on the false premise prompts. We attribute this to the nature of the prompts, i.e., providing additional contextual knowledge coerces the model to respond to a prompt even when the prompt is misleading; which increases the hallucination percentage.
Table \ref{tab:FPC_know_examples} shows examples of this result.
This is an important result because knowledge-augmented generation is typically considered to improve performance; however, we show that on false premise prompts, it instead proves to be detrimental. 
However, as expected, knowledge helps in answering the correct premise prompts as we show in the next study.

\paragraph{Impact of mitigation strategies on the correct premise prompts:}
Note that this study is crucial to highlight the negative impact of the mitigation strategies.
Figure \ref{fig:FPC_performance_both}(b) shows the performance of various mitigation strategies on the correct premise prompts.
Without any mitigation strategy (`Base'), the model correctly answers nearly all the instances. This is because the correct prompts are based on fundamental facts.
However, all strategies barring `Know' deteriorate the performance by hallucinating on the correct premise prompts.
This highlights an important downside of the mitigation strategies.
Unsurprisingly, `Know' does well on the correct premise prompts,
However, as noted before, it doesn't fair well on the false premise prompts where it coerces hallucination.
Interestingly, self-refinement also deteriorates the performance to a slight extent on the correct premise prompts. 
This is because during refinement, the model instead introduces hallucinations in the output.
We observe that in most of the deterioration cases, the model transformed the correct response by incorrectly introducing `not' into it.
Table \ref{tab:FPC_correct_premise_example_responses} shows examples of outputs of various strategies on the correct premise prompts.
Overall, `Inst + Exemp*' performs the best out of all the mitigation strategies as it reduces the hallucination on the false premise prompts while causing lower deterioration (relatively) on the correct premise prompts.

\subsection{Constrained Fact Generation}
\label{sec_results_CFG}

For both the fact generation tasks (CFG and FG), we evaluate the factual correctness of the model's output. 
For this evaluation, we use BARD (Gemini) model as it utilizes web search results to generate its output. 
Note that having web search access further assists in getting accurate evaluations for these tasks as they involve fact checking.
However, we also perform human annotations to measure the accuracy of BARD in evaluating the correctness and find it to be highly reliable (Appendix \ref{appendix_BARD_evaluation}).

\begin{table}[t]
    \resizebox{\linewidth}{!}{
    \begin{tabular}{p{3cm}|p{1.7cm}p{2cm}p{1.5cm}}
    \toprule
        \textbf{Models} & \textbf{LLaMA-2}  & \textbf{Vicuna-v1.5} &  \textbf{Orca-2} \\
    \toprule
        
        \textbf{Hallucination ($\downarrow$)} & 72\% & 73\% & 73\%\\

    \bottomrule
    \end{tabular}
    }
    \caption{
    Hallucination \% of models on the CFG task.}
    \label{tab:CFG_model_comparison}
\end{table}

Table \ref{tab:CFG_model_comparison} shows the hallucination percentage of models on this task. 
It shows that models falter significantly on this task as they generate a large percentage of hallucinated responses.
Table \ref{tab:CFG_example_responses} shows examples of responses of various models on this task.
We note that highlights a sycophantic behavior as the models tend to concoct a fact by simply combining the given keywords. 
This is a major limitation because it can potentially propagate misinformation.

\begin{table}[t]
    \resizebox{\linewidth}{!}{
    \begin{tabular}{p{1.5cm}|p{1.2cm}p{1.7cm}p{2cm}p{1.2cm}}
    \toprule
        \textbf{Models} & \textbf{Baseline} &  \textbf{LLaMA-2}  & \textbf{Vicuna-v1.5} &  \textbf{Orca-2} \\
    \toprule
        
        \textbf{Perf. ($\uparrow$)} & 51.4\%  & 62.2\%  & 54\% &  74\% \\

    \bottomrule
    \end{tabular}
    }
    \caption{
    Performance of models on the MCQA task.}
    \label{tab:MCQA_model_comparison}
\end{table}

\subsection{Multiple-Choice QA}
\label{sec_results_MCQA}

\paragraph{Performance Evaluation:}
In this task, we use the following performance metric:
\[
      \frac{c_r + i_{\neg r}}{\#\texttt{options}}
\]

where $c_r$ is the number of correct answer options in the response, $i_{\neg r}$ is the number of incorrect answer options not in the response, and \#\texttt{options} is the total number of answer options.

Table \ref{tab:MCQA_model_comparison} shows the performance of various models on this task. 
The table also shows a baseline system performance that corresponds to the system that includes all the answer options in its response; thus its performance equals to the number of correct options divided by the total number of options.
Orca-2 performs relatively better than other models on this task. This is because of its tuning methodology which is based on `explanation tuning', therefore, it explicitly tries to reason over all the options and then produces the final answer.
Table \ref{tab:MCQA_example_responses} shows examples of responses from Orca-2 on this task.
We also calculate the average number of answer options in the responses of all the models.
Specifically, LLaMA-2, Vicuna, and Orca-2 have $3.11$, $2.7$, and $3.84$ options in their respective responses and the average number of correct responses is $2.57$.

\subsection{Fact Generation}
\label{sec_results_FG}

Since LLMs are known to be brittle and sensitive to prompts, we experiment with three different prompts for this task.
Appendix \ref{appendix_sec_fg} provides all the prompts.
Furthermore, to compare models' ability to generate facts \textit{involving} and \textit{not involving} negation, we also generate facts using the following prompts:
(a) `Write five facts about \{{topic\}}. Each statement should be factually correct.'  
(b) `Write five accurate statements about \{{topic\}}.'
(c) `Share five true facts about \{{topic}\}.'

\begin{figure}[t!]
    \centering
    \includegraphics[width=7.5cm]{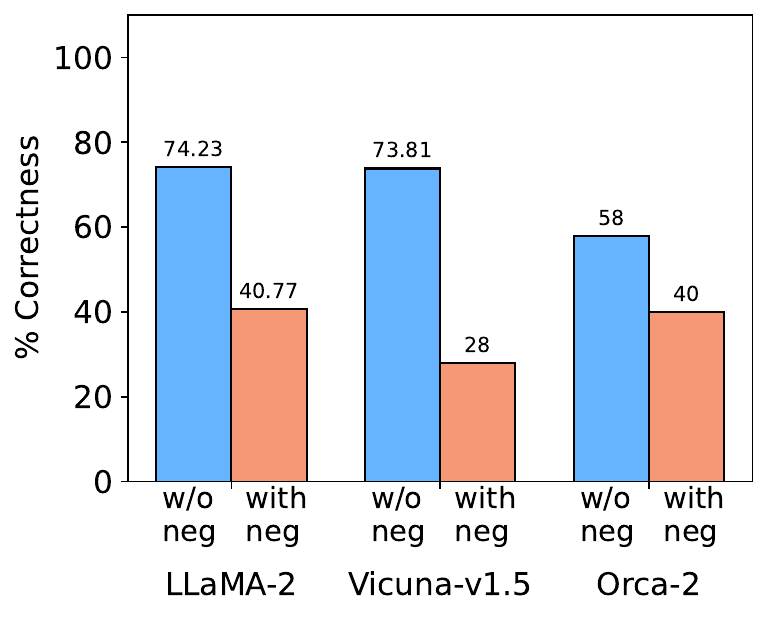}
    \caption{Performance of models on the FG task with negation (w/ neg) and without negation (w/o neg).}
    \label{fig:FG_models_comparison}
\end{figure}

\begin{figure}[t!]
    \centering
    \includegraphics[width=7.2cm]{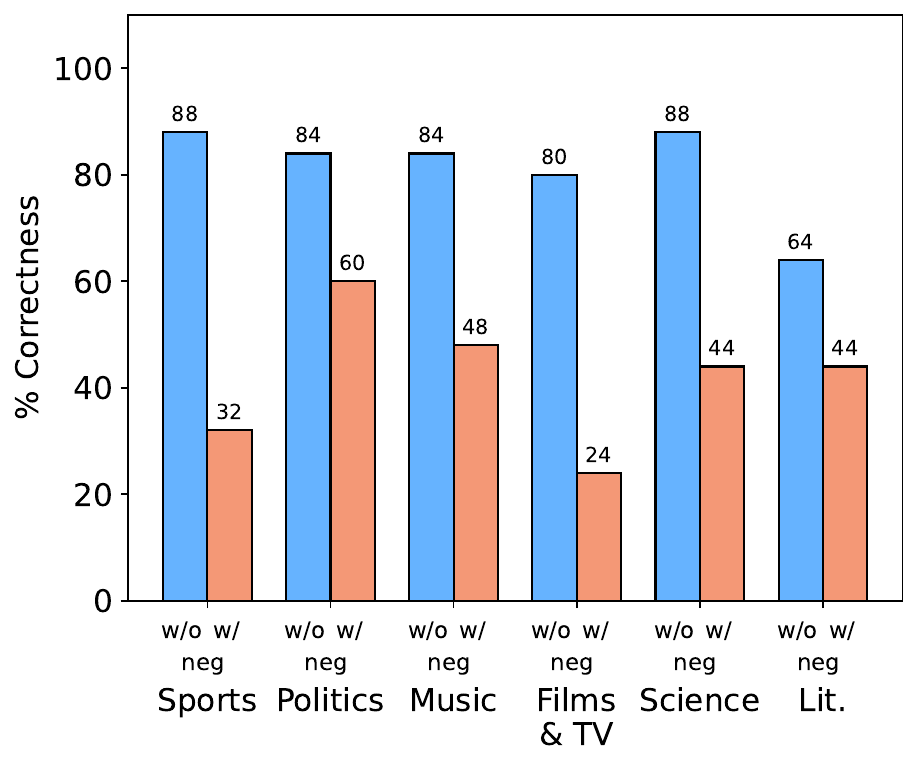}
    \caption{Domain-wise performance of LLaMA-2 on the FG task with negation and without negation.}
    \label{fig:FG_domains}
\end{figure}

Figure \ref{fig:FG_models_comparison} shows the performance of models for both `with negation' prompts (w/ neg) and `without negation' prompts (w/o neg).
On average, in the `w/o neg' setting, the hallucination percentage is $25.77\%$, $26.19\%$, and $42\%$ for the three models respectively while on the `w/ neg' scenario, the hallucination percentage increases to $59.23\%$, $72\%$, and $60\%$ for the three models.
This shows the models hallucinate considerably higher in generating facts containing negation.
We further show this comparison on each domain for the LLaMA-2 model in Figure \ref{fig:FG_domains}.
The same finding holds true across all the domains.
Table \ref{tab:examples_international} shows examples of facts generated for both `without negation' and `with negation' prompts.

\section{Conclusion}
Motivated by the importance of `negation' in language, we comprehensively studied hallucinations of LLMs with four evaluation tasks that deal with negation.
Specifically, we studied false premise completion, constrained fact generation, multiple-choice QA, and fact generation tasks.
Through extensive experiments,  we demonstrated that open-source state-of-the-art LLMs including LLaMA-2-chat, Vicuna-v1.5, and Orca-2 hallucinate considerably on all these tasks.
This highlights a critical shortcoming of these models.
Addressing this hallucination problem, we also studied various mitigation strategies and analyzed their effectiveness. 
Our analysis also revealed an important negative impact of employing these mitigation strategies.
In summary, our work underlined a critical shortcoming of existing LLMs pertaining to their ability to deal with negation and studied ways to address that limitation.

\section*{Limitations}

Like any other aspect of language, `negation' has applicability in a variety of tasks.
In this work, we make an attempt to capture this aspect by studying four distinct tasks, namely, false premise completion, constrained fact generation, multiple-choice QA, and fact generation.
However, it is not an exhaustive list covering the entire landscape of negation.
In the future, the study can be extended by including more tasks that deal with negation.
More and more large language models are being developed at a rapid pace; however, in this work, we have only considered widely used LLMs. As more models get developed and gain prominence, the study can be easily extended to incorporate their results.
Finally, our dataset includes questions in only one language, i.e., English. Investigating the multilingual setting is out of the scope of this work but can be pursued by follow-up works in future.

\section*{Ethical Considerations}

For all the tasks, we have only considered widely known facts/personalities to avoid any sort of bias.
The examples do not contain any private, confidential, or controversial information.
We have used AI assistants (Grammarly and ChatGPT) to address the grammatical errors and rephrase the sentences.

\bibliography{anthology,custom}
\bibliographystyle{acl_natbib}

\appendix
\section*{Appendix}

\section{Additional Experimental Details}

\subsection{Demonstrative Exemplars (Exemp)}
\label{appendix_exemp}

LLMs possess an impressive ability to learn from demonstrations/exemplars provided in the input context.
To this end, we provide exemplars of input-output pairs of negation prompts in the context.
Each exemplar is in the form of (input, output).
We use different combinations of the following exemplars:

(a) Input: Neptune is not wider than Earth because

Output: I must correct the statement: Neptune is indeed wider than Earth. Neptune is the eighth and farthest known planet from the Sun in our solar system, and it is a gas giant with a diameter of about 49,244 kilometers (30,598 miles).,

(b) Input: Australia did not win the 2007 cricket world cup because

Output: Australia did indeed win the 2007 Cricket World Cup. They defeated Sri Lanka in the final, held on April 28, 2007, at the Kensington Oval in Barbados.,

(c) Input: Rabindranath Tagore did not receive the Nobel Prize for Literature because

Output: Rabindranath Tagore did, in fact, receive the Nobel Prize for Literature. He was awarded the prestigious prize in 1913, making him the first Asian to be honored with a Nobel Prize.,

(d) Input: Neptune is not narrower than Earth because

Output: Neptune is not narrower than Earth because of fundamental differences in their compositions, sizes, and positions in the solar system. It's essential to clarify that Neptune is not narrower than Earth; in fact, it is significantly larger.,
    
(e) Input: Australia did not win the 2011 cricket world cup because

Output: Australia did not win the 2011 Cricket World Cup because they were eliminated in the quarterfinals of the tournament.,

(f) Input: Rabindranath Tagore did not receive the Nobel Prize for Physics because

Output: Rabindranath Tagore did not receive the Nobel Prize for Physics because his Nobel Prize, awarded in 1913, was specifically for Literature. Tagore was honored with the Nobel Prize in Literature for his collection of poems titled 'Gitanjali',

(g) Input: Neptune is narrower than Earth because

Output: I must correct the statement: Neptune is indeed wider than Earth. Neptune is the eighth and farthest known planet from the Sun in our solar system, and it is a gas giant with a diameter of about 49,244 kilometers (30,598 miles).,

(h) Input: Australia won the 2011 cricket world cup because

Output: Australia did not win the 2011 Cricket World Cup. The 2011 ICC Cricket World Cup was won by India.,

(i) Input: Rabindranath Tagore received the Nobel Prize for Physics because

Output: Rabindranath Tagore did not receive the Nobel Prize for Physics. He received the Nobel Prize for Literature. He was awarded the prestigious prize in 1913, making him the first Asian to be honored with a Nobel Prize.,

(j) Input: Neptune is wider than Earth because

Output: Neptune is indeed wider than Earth primarily because it is a larger planet in terms of both diameter and volume.,

(k) Input: Australia won the 2007 cricket world cup because

Output: Australia won the 2007 Cricket World Cup primarily due to their exceptional team performance throughout the tournament.,

(l) Input: Rabindranath Tagore received the Nobel Prize for Literature because

Output: He received this prestigious honor primarily for his collection of poems titled 'Gitanjali' (Song Offerings).,

For `Inst + Exemp' strategy, we experiment with three different combinations of exemplars: [a,b,d,e], [a,c,d,f], and [a,b,e,f] and report averaged results.

For `Inst + Exemp*' strategy, we experiment with the following examples [a,b,d,e,g,h,j,k].

\subsection{Knowledge Augmentation (Know):}
\label{appendix_know}

Table \ref{tab:FPC_know_examples} shows examples of knowledge retrieved for various prompts.
We use snippets returned by Bing Search API of two search results as knowledge.

\begin{table*}[t]
    \small
    \centering
    \resizebox{\linewidth}{!}{
    \begin{tabular}{@{}p{0.18\linewidth}>{\RaggedRight}p{0.8\linewidth}@{}}
    \toprule
        \textbf{Prompt } &
        \textbf{Knowledge} 
        \\
    \toprule

    Jupiter is not bigger than Earth because

    & Jupiter: Facts - NASA Science, Quick Facts Eleven Earths could fit across Jupiter’s equator. If Earth were the size of a grape, Jupiter would be the size of a basketball. Jupiter orbits about 484 million miles (778 million kilometers) or 5.2 Astronomical Units (AU) from our Sun (Earth is one AU from the Sun)...

    Jupiter - Wikipedia, Formation and migration Jupiter is believed to be the oldest planet in the Solar System, having formed just one million years after the Sun and roughly 50 million years before Earth. [23] ... \\

    \midrule

    Metals are not a good conductor of heat because

    & 7.6: Metals, Nonmetals, and Metalloids - Chemistry LibreTexts, Valency: Metals typically have 1 to 3 electrons in the outermost shell of their atoms. Conduction: Metals are good conductors because they have free electrons. Silver and copper are the two best conductors of heat and electricity. Lead is the poorest conductor of heat. Bismuth, mercury and iron are also poor conductors ...
    
    2.11: Metals, Nonmetals, and Metalloids - Chemistry LibreTexts, Conduction: Metals are good conductors because they have free electrons. Silver and copper are the two best conductors of heat and electricity. Lead is the poorest conductor of heat. Bismuth, mercury and iron are also poor conductors; Density: Metals have high density and are very heavy. Iridium and osmium have the highest densities where as ... \\

    \bottomrule

    \end{tabular}
    }
    \caption{Examples of knowledge retrieved by using the corresponding prompt as the search query.}
    \label{tab:FPC_know_examples}
\end{table*}

\section{False Premise Completion}
\label{appendix_sec_fpc}

\begin{table*}[t]
    \small
    \centering
    \resizebox{\linewidth}{!}{
    \begin{tabular}{@{}p{0.18\linewidth}>{\RaggedRight}p{0.8\linewidth}@{}}
    \toprule
        \textbf{Prompt Type/Model} &
        \textbf{Prompt/Model Response} 
        \\
    \toprule



    


    \textbf{False Premise Prompt} & Sound does not travel through air because \\


    \textbf{LLaMA-2} & Sound does not travel through air because air is a poor conductor of sound waves. Sound waves ... (\xmark) \\

    \textbf{Vicuna} & Sound does not travel through air because sound waves are mechanical waves that require a medium to propagate. In the case of air, sound waves cannot travel because air is an incompressible fluid, meaning it does not have the ability to vibrate and transmit sound waves. (\xmark) \\
    
    \textbf{Orca-2} & Sound does not travel through air because it is a mechanical wave that requires a medium to propagate. In the case of air, it is not dense enough to transfer sound waves efficiently. (\xmark) \\

    \midrule

    \textbf{False Premise Prompt} & Metals are not a good conductor of heat because \\


    \textbf{LLaMA-2} & Metals are not a good conductor of heat because they have a low thermal conductivity. Thermal conductivity is a measure of how easily  ... (\xmark) \\

    \textbf{Vicuna} & Metals are not a good conductor of heat because their atoms are not loosely packed, which makes it difficult for heat to travel through them. ... (\xmark) \\
    
    \textbf{Orca-2} & Metals are not a good conductor of heat because they have a low thermal conductivity. This means that they do not transfer heat efficiently from one part to another. The atoms in metals are closely packed together, which makes it difficult for heat to move through them ... (\xmark) \\

    \midrule

    \textbf{False Premise Prompt} & The speed of sound is not affected by the medium through which it travels because \\


    \textbf{LLaMA-2} & Sorry, it's not accurate to say that the speed of sound is not affected by the medium through which it travels. In fact, the speed of sound can vary depending on the medium it is traveling through.
    The speed of sound is affected by the properties of the medium it is traveling through, such as its density, elasticity, and temperature.  (\cmark) \\

    \textbf{Vicuna} & The speed of sound is not affected by the medium through which it travels because the speed of sound is a fundamental physical constant, and it is the same in all media. ... (\xmark) \\
    
    \textbf{Orca-2} & The speed of sound is not affected by the medium through which it travels because it is determined by the properties of the sound source and the receiving listener's ears, rather than the medium itself. ... (\xmark) \\

    \bottomrule

    \end{tabular}
    }
    \caption{Examples of responses of various models on the false premise prompts.}
    \label{tab:FPC_example_responses}
\end{table*}

\begin{table*}[t]
    \small
    \centering
    \resizebox{\linewidth}{!}{
    \begin{tabular}{@{}p{0.48\linewidth}>{\RaggedRight}p{0.48\linewidth}@{}}
    \toprule
        \textbf{False Premise Prompt } &
        \textbf{Correct Premise Prompt} 
        \\
    \toprule

    Saturn is not the second largest planet in our solar system because &  Saturn is the second largest planet in our solar system because\\

    Jupiter is not bigger than Earth because & Jupiter is bigger than Earth because\\
    
    Golf balls are not smaller than basketballs because & Golf balls are smaller than basketballs because\\

    Metals are not a good conductor of heat because & Metals are a good conductor of heat because\\

    Tennis balls do not have a smaller diameter than soccer balls because & Tennis balls have a smaller diameter than soccer balls because\\

    Heat energy does not transfer from a warmer substance to a colder one because & Heat energy transfers from a warmer substance to a colder one because\\

    Hydrogen does not have atomic number of 1 because & Hydrogen has atomic number of 1 because\\

    Inflation does not decrease the purchasing power of money because & Inflation decreases the purchasing power of money because\\

    Dolphins are not mammals because & Dolphins are mammals because\\

    \bottomrule

    \end{tabular}
    }
    \caption{Examples of false premise prompts with corresponding correct premise prompts.}
    \label{tab:FPC_correct_premise_prompts}
\end{table*}

Table \ref{tab:FPC_example_responses} shows examples of responses of various models on the false premise prompts. 

Table \ref{tab:FPC_correct_premise_prompts} shows examples of false premise prompts and corresponding correct premise prompts.

\begin{table*}[t]
    \small
    \centering
    \resizebox{\linewidth}{!}{
    \begin{tabular}{@{}p{0.18\linewidth}>{\RaggedRight}p{0.8\linewidth}@{}}
    \toprule
        \textbf{Prompt/Strategy} &
        \textbf{Prompt/Model Response} 
        \\
    \toprule

    \textbf{Prompt} & Saturn is not the second largest planet in our solar system because \\


    \textbf{Base} & Saturn is not the second largest planet in our solar system because it is actually the sixth largest planet in our solar system. (\xmark) \\

    \textbf{Self-Refinement} & Saturn is not the second largest planet in our solar system because it is actually the sixth largest planet in our solar system. (\xmark) \\
    
    \textbf{Inst + Exemp} & Saturn is, in fact, the second largest planet in our solar system, with a diameter of approximately 116,464 kilometers (72,337 miles). The largest planet in our solar system is Jupiter, which has a diameter of around ...(\cmark)\\

    \midrule

    \textbf{Prompt} & Golf balls are not smaller than basketballs because \\


    \textbf{Base} & Golf balls are not smaller than basketballs because golf balls are actually larger in diameter than basketballs. A standard golf ball is approximately 1.68 inches (4.27 cm) in diameter, while a standard basketball is approximately 1.21 inches (3.07 cm) in diameter. So, golf balls are about 25\% larger in diameter than basketball (\xmark) \\

    \textbf{Self-Refinement} &  Golf balls are not larger than basketballs because golf balls are actually smaller in diameter than basketballs. (\cmark) \\
    
    \textbf{Inst } & Contrary to popular belief, golf balls are actually larger than basketballs.  (\xmark)\\

    \textbf{Inst + Exemp*} & Golf balls are actually larger than basketballs. A standard golf ball is about 4.2 cm (1.65 inches) in diameter, while a standard basketball is about 2.5 cm (0.98 inches) in diameter. (\xmark)\\

    \midrule

    \textbf{Prompt} & Chickens do not lay eggs because \\

    \textbf{Base} & Chickens do not lay eggs because they are mammals, not birds. Chickens, like all other mammals, give birth to live young instead of laying eggs. This is why you will never find eggs in a chicken coop or from a chicken farm.  (\xmark) \\

    \textbf{Self-Refinement} &  Chickens do not lay eggs because they are birds, not mammals. Chickens, like all other birds, lay eggs instead of giving birth to live young. This is why you will often find eggs in a chicken coop or from a chicken farm. (\xmark) \\
    
    \textbf{Inst } & Chickens do not lay eggs because they are mammals, not birds. Chickens, like all other mammals, give birth to live young instead of laying eggs.  (\xmark)\\

    \textbf{Inst + Exemp} & Chickens do lay eggs! In fact, chickens are one of the most common sources of eggs for human consumption. Hens, the female chickens, lay eggs as part of their reproductive cycle, and the eggs are typically collected from farms and sold in grocery stores (\cmark)\\

    \bottomrule

    \end{tabular}
    }
    \caption{Examples of responses after application of various mitigation strategies on the false premise prompts.}
    \label{tab:FPC_flase_premise_mitigation_responses}
\end{table*}

Table \ref{tab:FPC_flase_premise_mitigation_responses} shows examples of responses after application of various mitigation strategies on the false premise prompts.

Table \ref{tab:FPC_knowledge_examples} shows examples of responses after application of `Know' mitigation strategy on the false premise prompts.

\begin{table*}[t]
    \small
    \centering
    \resizebox{\linewidth}{!}{
    \begin{tabular}{@{}p{0.18\linewidth}>{\RaggedRight}p{0.8\linewidth}@{}}
    \toprule
        \textbf{Prompt/Strategy} &
        \textbf{Prompt/Model Response} 
        \\
    \toprule

    \textbf{Prompt} & The Sun does not emit light because \\

    \textbf{Base} & the statement ``The Sun does not emit light because'' is not entirely accurate. The Sun does emit light, and in fact, it is the primary source of light and energy for our solar system. The Sun emits light through a process called nuclear fusion, where hydrogen atoms are fused together to form helium, releasing a vast amount of energy in the form of light and heat. This process occurs at the Sun's core. (\cmark) \\

    \textbf{Know} & The Sun does not emit light because it is an ordinary star and does not have the ability to produce light on its own.  (\xmark) \\

    \midrule

    \textbf{Prompt} & The speed of sound is not affected by the medium through which it travels because \\

    \textbf{Base} & Sorry, it's not accurate to say that the speed of sound is not affected by the medium through which it travels. In fact, the speed of sound can vary depending on the medium it is traveling through.
    
    The speed of sound is affected by the properties of the medium it is traveling through, such as its density, elasticity, and temperature. For example, sound travels faster through solids than through liquids, and faster through liquids than through (\cmark) \\

    \textbf{Know} & The speed of sound is not affected by the medium through which it travels because the speed of sound is a fundamental physical constant, and it remains constant regardless of the medium through which it travels. (\xmark) \\

    \bottomrule

    \end{tabular}
    }
    \caption{Examples of responses after application of `Know' mitigation strategy on the false premise prompts.}
    \label{tab:FPC_knowledge_examples}
\end{table*}

\begin{table*}[t]
    \small
    \centering
    \resizebox{\linewidth}{!}{
    \begin{tabular}{@{}p{0.18\linewidth}>{\RaggedRight}p{0.8\linewidth}@{}}
    \toprule
        \textbf{Prompt/Strategy} &
        \textbf{Prompt/Model Response} 
        \\
    \toprule

    \textbf{Prompt} & Golf balls are smaller than basketballs because \\


    \textbf{Base} & Golf balls are smaller than basketballs because golf balls are designed to be smaller and lighter to allow for longer distances and more accurate shots, while basketballs are larger and heavier to provide more stability and control during play.(\cmark) \\

    \textbf{Self-Refinement} & Golf balls are not smaller than basketballs. In fact, golf balls are slightly larger than basketballs. This is because golf balls are designed to be more aerodynamic and have a higher lift ...(\xmark) \\
    
    \textbf{Know} & Golf balls are smaller than basketballs because golf balls are regulation size 5 balls, which have a circumference of 68cm to 70.5cm and weigh between 445g and 475g when fully inflated. In comparison, basketballs are typically size 7 or size 8, ... (\cmark)\\

    \midrule 
    
     \textbf{Prompt} & Snakes are reptiles because \\

    \textbf{Base} & Snakes are reptiles because they have several characteristics that are common to all reptiles, such as:
 
    1. Cold-bloodedness: Snakes, like all reptiles, are ectothermic, meaning their body temperature is regulated by the environment around them, rather than being able to generate heat internally ... (\cmark) \\

    \textbf{Inst + Exemp} & Snakes are not reptiles because they do not have legs or external ears, which are characteristics of reptiles ... (\xmark) \\
    
    \textbf{Self-Refinement} & Snakes are not reptiles because they do not have several characteristics that are common to all reptiles, such as:
     1. Cold-bloodedness: Snakes, unlike all reptiles, are not ectothermic, meaning their body temperature is not regulated by the environment around them, but rather internally. ... (\xmark)\\

     \midrule

    \textbf{Prompt} & Ice is indeed less dense than water \\

    \textbf{Base} &  Ice is less dense than water because of the way the molecules are arranged in each state of matter. In its solid form, ice has a crystalline structure that is made up of rigid molecules ... (\cmark) \\

    \textbf{Inst} & I think there may be a mistake in the prompt. Ice is actually more dense than water, not less dense. (\xmark) \\
    
    \textbf{Self-Refinement} &  Ice is less dense than liquid water because of the way the molecules are arranged in each state of matter. In its solid form, ice has a crystalline structure that is made up of rigid molecules  ... (\cmark)\\

    \bottomrule

    \end{tabular}
    }
    \caption{Examples of responses after application of various mitigation strategies on the correct premise prompts.}
    \label{tab:FPC_correct_premise_example_responses}
\end{table*}

Table \ref{tab:FPC_correct_premise_example_responses} shows examples of responses after application of various mitigation strategies on the correct premise prompts.

We also that it is also plausible to correctly complete the negated prompt by providing correct supporting details. 
Consider another prompt, ``Trees do not produce oxygen because'', a response mentioning ``Trees do not produce the same amount of oxygen year-round because oxygen production is dependent on sunlight and other factors'' is correct; in contrast, a response ``Trees do not produce because they instead produce nitrogen during photosynthesis'' is incorrect.

\paragraph{Self-Checking Analysis}
For further analysis, we additionally evaluate the ability of the models to correctly check the factuality of the prompts. 
To this end, we experiment with the following three different variants of the input: 
``Is the given prompt factually correct?'', 
``Does the provided prompt contain factually accurate information?'', and 
``Is the information presented in the prompt factually true?''.

We provide both false premise and correct premise prompts as input. 
The averaged accuracy of the LLaMA model on this task is $62.7\&$ just slightly above the random baseline. 
This shows the limitation of the model in self-checking the factuality of the prompt.

\section{Constrained Fact Generation}
\label{appendix_sec_cfg}

\begin{table*}[t]
    \small
    \centering
    \resizebox{\linewidth}{!}{
    \begin{tabular}{@{}p{0.24\linewidth}>{\RaggedRight}p{0.76\linewidth}@{}}
    \toprule
        \textbf{Input/Model} &
        \textbf{Keywords/Model Response} 
        \\
    \toprule

    \textbf{Keywords} & Michael Clarke, Cricket World Cup, not, 2015 \\

    \textbf{LLaMA-2} & Michael Clarke did not play in the 2015 Cricket World Cup. (\xmark) \\

    \textbf{Vicuna} & Michael Clarke did not participate in the 2015 Cricket World Cup. (\xmark) \\
    
    \textbf{Orca-2} & Michael Clarke did not play in the 2015 Cricket World Cup due to a hamstring injury. (\xmark) \\

    \textbf{Possible Correct Generation} & Michael Clarke did not score a century in the 2015 Cricket World Cup. (\cmark) \\

    \midrule

    \textbf{Keywords} & not, Finland, Men's Ice Hockey World Championships \\

    \textbf{LLaMA-2} & Finland has never won the Men's Ice Hockey World Championships, despite being one of the top teams in the tournament's history.  (\xmark) \\

    \textbf{Vicuna} & Finland did not win the Men's Ice Hockey World Championships in 2021.  (\cmark) \\
    
    \textbf{Orca-2} & Finland did not qualify for the Men's Ice Hockey World Championships in 2021.  (\xmark) \\

    \bottomrule

    \end{tabular}
    }
    \caption{Examples of responses of various models on the CFG task.}
    \label{tab:CFG_example_responses}
\end{table*}

Table \ref{tab:CFG_example_responses} shows examples of responses of various models on the CFG task.
Though the scope of this project is limited to open-source 13B models, we also evaluate GPT-4 model on this task and found that even GPT-4 hallucinates on $60\%$ instances.
This result further opens up avenues for future research in evaluating and improving the ability of LLMs to effectively deal with negation.
up several avenues for new research
Table \ref{tab:CFG_GPT4_examples} provides examples of GPT-4 responses.

\begin{table*}[t]
    \small
    \centering
    \resizebox{\linewidth}{!}{
    \begin{tabular}{@{}p{0.24\linewidth}>{\RaggedRight}p{0.76\linewidth}@{}}
    \toprule
        \textbf{Input/Model} &
        \textbf{Keywords/Model Response} 
        \\
    \toprule

    \textbf{Keywords} & 8, not, Oxygen, Atomic Number \\

    \textbf{GPT-4} & The atomic number of Oxygen is not 8. (\xmark) \\

    \midrule

    \textbf{Keywords} & Roger Federer, not, Wimbledon \\

    \textbf{GPT-4} & Roger Federer has not won Wimbledon since 2017.(\cmark) \\

    \midrule

    \textbf{Keywords} & 2016, Denver Broncos, Super Bowl, Not \\

    \textbf{GPT-4} & In 2016, the Denver Broncos did not lose the Super Bowl; they won it.(\cmark) \\

    \midrule

    \textbf{Keywords} & Claire Foy, not, emmy awards, 2018 \\

    \textbf{GPT-4} & Claire Foy did not win an Emmy Award in 2018.(\xmark) \\

    \bottomrule

    \end{tabular}
    }
    \caption{Examples of responses of GPT-4 on the CFG task.}
    \label{tab:CFG_GPT4_examples}
\end{table*}

\section{Multiple-Choice QA}
\label{appendix_sec_mcqa}

\begin{table*}[t]
    \small
    \centering
    \resizebox{\linewidth}{!}{
    \begin{tabular}{@{}p{0.14\linewidth}>{\RaggedRight}p{0.85\linewidth}@{}}
    \toprule

    \textbf{Question} & Identify the Formula 1 drivers who have not won the Monaco Grand Prix. \\

    \textbf{Options} & (A)  Max Verstappen, 	(B)  Ayrton Senna, 	\textbf{(C) Carlos Sainz}, 	(D) Lewis Hamilton, 	\textbf{(E)  Lando Norris} \\ \\


    \multirow{12}{*}{\textbf{Orca-2 Response}} & The question is asking for the Formula 1 drivers who have not won the Monaco Grand Prix. The options provided are Max Verstappen, Ayrton Senna, Carlos Sainz, Lewis Hamilton, and Lando Norris.

First, let's eliminate the drivers who we know have won the Monaco Grand Prix. Ayrton Senna and Lewis Hamilton are both well-known for their success in Formula 1, including multiple Monaco Grand Prix wins. Therefore, we can eliminate options (B) and (D).

Now, let's consider the remaining options: Max Verstappen, Carlos Sainz, and Lando Norris. All three of these drivers are currently active in Formula 1, and it is possible that they have not yet won the Monaco Grand Prix. However, Max Verstappen has won the Monaco Grand Prix in the past, so we can eliminate option (A).

This leaves us with two options: Carlos Sainz and Lando Norris. Both of these drivers are still active in Formula 1, and as of the time of this writing, neither of them has won the Monaco Grand Prix. Therefore, both options (C) and (E) are correct.

\#\#\# Final Answer: (C) Carlos Sainz, (E) Lando Norris \\

\midrule

 \textbf{Question} & Select all cricketers who were not part of the 2007 world cup winning team. \\

    \textbf{Options} & \textbf{(A)   Ryan Harris}, 	(B) Shaun Tait, 	\textbf{(C) Brett Lee}, 	\textbf{(D)   Shane Warne}, 	(E)  Brad Hogg \\ \\


    \multirow{10}{*}{\textbf{Orca-2 Response}} & 
    Key concepts and information:
    - Cricketers
    - 2007 world cup winning team
    
    Breakdown of the question:
    - Identify which cricketers were not part of the 2007 world cup winning team
    
    Relevant information:
    - The 2007 world cup winning team was Australia
    
    Answer choices:
    (A) Ryan Harris
    (B) Shaun Tait
    (C) Brett Lee
    (D) Shane Warne
    (E) Brad Hogg
    
    Elimination of answer choices:
    - Shane Warne was part of the 2007 world cup winning team, so eliminate (D)
    
    Narrowing down choices:
    - We are left with (A) Ryan Harris, (B) Shaun Tait, (C) Brett Lee, and (E) Brad Hogg
    
    Comparison of remaining choices:
    - All four remaining choices were not part of the 2007 world cup winning team
    
    Final Answer: (A) Ryan Harris, (B) Shaun Tait, (C) Brett Lee, (E) Brad Hogg\\

    \bottomrule

    \end{tabular}
    }
    \caption{Examples of Orca-2 responses on the MCQA task.}
    \label{tab:MCQA_example_responses}
\end{table*}

Orca-2 performs relatively better than other models on this task. This is because of its tuning methodology which is based on `explanation tuning', therefore, it explicitly tries to reason over all the options and then produces the final answer.
Table \ref{tab:MCQA_example_responses} shows examples of responses from Orca-2 on the MCQA task.
\section{Fact Generation}
\label{appendix_sec_fg}

\begin{table*}[t]
    \small
    \centering
    \resizebox{\linewidth}{!}{
    \begin{tabular}{@{}p{0.13\linewidth}>{\RaggedRight}p{0.4\linewidth}>{\RaggedRight}p{0.4\linewidth}}
    \toprule
        \textbf{Name} &
        \textbf{Without Negation} &
        \textbf{With Negation}
        \\
    \midrule

    \multirow{10}{*}{Virat Kohli}
    & Kohli has won numerous awards for his cricketing achievements, including the ICC Cricketer of the Year award in 2017 and 2018. (\cmark) & Kohli has never scored a double century in Test cricket, despite having a career average of over 50 in the format. (\xmark)\\

    &  He is a right-handed batsman and has been ranked as the number one Test batsman in the world by the International Cricket Council (ICC) on several occasions. (\cmark) & 4. Virat Kohli has never won the ICC Cricketer of the Year award, although he has been nominated several times. (Negation: has never won) (\xmark)\\

    & He has scored over 11,000 runs in One Day Internationals (ODIs) and has a batting average of over 50 in this format. (\cmark) & Virat Kohli does not have a perfect record in Test cricket, having been dismissed for a duck (zero) on several occasions.(\cmark) \\

    \bottomrule

    \end{tabular}
    }
    \caption{Examples of facts generated by LLaMA-2 for both `without negation' and `with negation' prompt. (\xmark) and (\cmark) correspond to hallucinated and factually correct statements respectively.}
    \label{tab:examples_international}
\end{table*}

Since LLMs are known to be brittle and sensitive to prompts, we experiment with three different prompts for this task: 
(a) `{Write five statements about \{{topic\}}, each containing a negation and presenting factual information.}' 
(b) `{Compose five factually accurate statements that include negation related to \{{topic\}}.}' 
(c) `{Write five factual statements about \{{topic\}}, all of which include negation.}' 

Furthermore, to compare models' ability to generate facts involving and not involving negation, we also generate facts using the following prompts:
(a) `Write five facts about \{{topic\}}. Each statement should be factually correct.'  
(b) `Write five accurate statements about \{{topic\}}.'
(c) `Share five true facts about \{{topic}\}.'

Table \ref{tab:examples_international} shows examples of facts generated by LLaMA-2 for both `without negation' and `with negation' prompt.

\section{BARD Evaluations}
\label{appendix_BARD_evaluation}

For both the fact generation tasks (CFG and FG), we evaluate the factual correctness of the model's output. 
For this evaluation, we use BARD (Gemini) model as it utilizes web search results to generate its output. 
Note that having web search access further assists in getting accurate evaluations for these tasks as they involve checking factual correctness.
Furthermore, BARD performs remarkably well in providing nuanced explanations while verifying the facts. 
It is able to dissect and consider every aspect of the response to construct the overall evaluation of the correctness of the statements.
However, we also perform human annotations to measure the accuracy of BARD in evaluating the correctness.
Table \ref{tab:bard_evaluation_examples} shows examples of responses given by BARD.

\begin{table*}[t]
    \small
    \centering
    \resizebox{\linewidth}{!}{
    \begin{tabular}{@{}p{0.25\linewidth}>{\RaggedRight}p{0.74\linewidth}@{}}
    \toprule
        \textbf{Text } &
        \textbf{BARD Response} 
        \\
    \toprule

    Cristiano Ronaldo was born on February 5, 1985, in Funchal, Madeira, Portugal.

    & 
This statement is correct.

Here's why:

    Multiple credible sources report Cristiano Ronaldo's birth date and place as February 5, 1985, in Funchal, Madeira, Portugal. These sources include:
        Wikipedia: https://en.wikipedia.org/wiki/Cristiano\_Ronaldo
        Britannica: https://www.britannica.com/biography/Cristiano-Ronaldo
        Biography: https://www.biography.com/athletes/cristiano-ronaldo
    Ronaldo himself has confirmed his birth date and place in interviews and on social media.

Therefore, based on the available evidence, the claim that Cristiano Ronaldo was born on February 5, 1985, in Funchal, Madeira, Portugal, is correct. \\

   \midrule

   Cristiano Ronaldo has never represented Portugal in a World Cup tournament. & 

   The statement that Cristiano Ronaldo has never represented Portugal in a World Cup tournament is incorrect. He has actually represented Portugal in five World Cups: 2006, 2010, 2014, 2018, and 2022. In fact, he is the all-time leading scorer for Portugal in the World Cup with eight goals. \\

   \midrule

 Despite being a prolific scorer, Ronaldo has never won the Golden Boot award at the UEFA Champions League, although he has come close several times. &

 Incorrect.

Explanation:

Cristiano Ronaldo has actually won the UEFA Champions League Golden Boot award a record seven times, not never. He holds the distinction of being the top scorer in the competition the most times, surpassing his long-time rival Lionel Messi who has won it six times. Here's a breakdown of Ronaldo's Champions League Golden Boot wins:

2007-08: Manchester United
2013-14: Real Madrid (17 goals, record for most goals in a single season)
2014-15: Real Madrid (tied with Neymar and Messi, all at 10 goals)
2015-16: Real Madrid
2016-17: Real Madrid
2017-18: Real Madrid
2018-19: Juventus
Therefore, the claim that Ronaldo has never won the Champions League Golden Boot is factually incorrect. \\
        
    \bottomrule

    \end{tabular}
    }
    \caption{Examples of BARD evaluations.}
    \label{tab:bard_evaluation_examples}
\end{table*}

\end{document}